\documentclass[10pt,twocolumn,letterpaper]{article}
\pdfoutput=1  

\usepackage[pagenumbers]{cvpr} 
\usepackage{graphicx}
\usepackage{caption}
\usepackage{subcaption}  
\usepackage{placeins}

\usepackage{amsmath,amsfonts}
\usepackage{graphicx}
\usepackage{booktabs}
\usepackage{subcaption}
\usepackage{siunitx}
\usepackage{tikz}
\usepackage{pgfplots}
\usepackage{pgfplotstable}
\usetikzlibrary{shapes,arrows,positioning,fit,calc}
\pgfplotsset{compat=1.18}

\tikzstyle{block} = [rectangle, draw, fill=blue!10, text centered, rounded corners, minimum height=2.5em, minimum width=6em, text width=8em]
\tikzstyle{fblock} = [rectangle, draw, fill=green!10, text centered, rounded corners, minimum height=2.5em, minimum width=6em, text width=8em]
\tikzstyle{connector} = [draw, -latex']
\tikzstyle{line} = [draw, thick]
\tikzstyle{cloud} = [ellipse, draw, fill=red!10, minimum height=2em]
\tikzstyle{multi} = [rectangle, draw, fill=gray!10, text centered, rounded corners, minimum height=2em]

\definecolor{cvprblue}{rgb}{0.21,0.49,0.74}
\usepackage[pagebackref,breaklinks,colorlinks,allcolors=cvprblue]{hyperref}

\def\paperID{*****} 
\def\confName{CVPR}
\def\confYear{2026}

\title{DARN: Dynamic Adaptive Regularization Networks for Efficient and Robust Foundation Model Adaptation}

\author{
Dhenenjay Yadav\\
AxionOrbital Space\\
{\tt\small dhenenjay@axionorbital.space}
\and
Rohan Sawai\\
Virginia Tech\\
{\tt\small sawairohan90@vt.edu}
}

\begin{document}
\maketitle

\begin{figure*}[!ht]
    \centering
    \includegraphics[width=\textwidth]{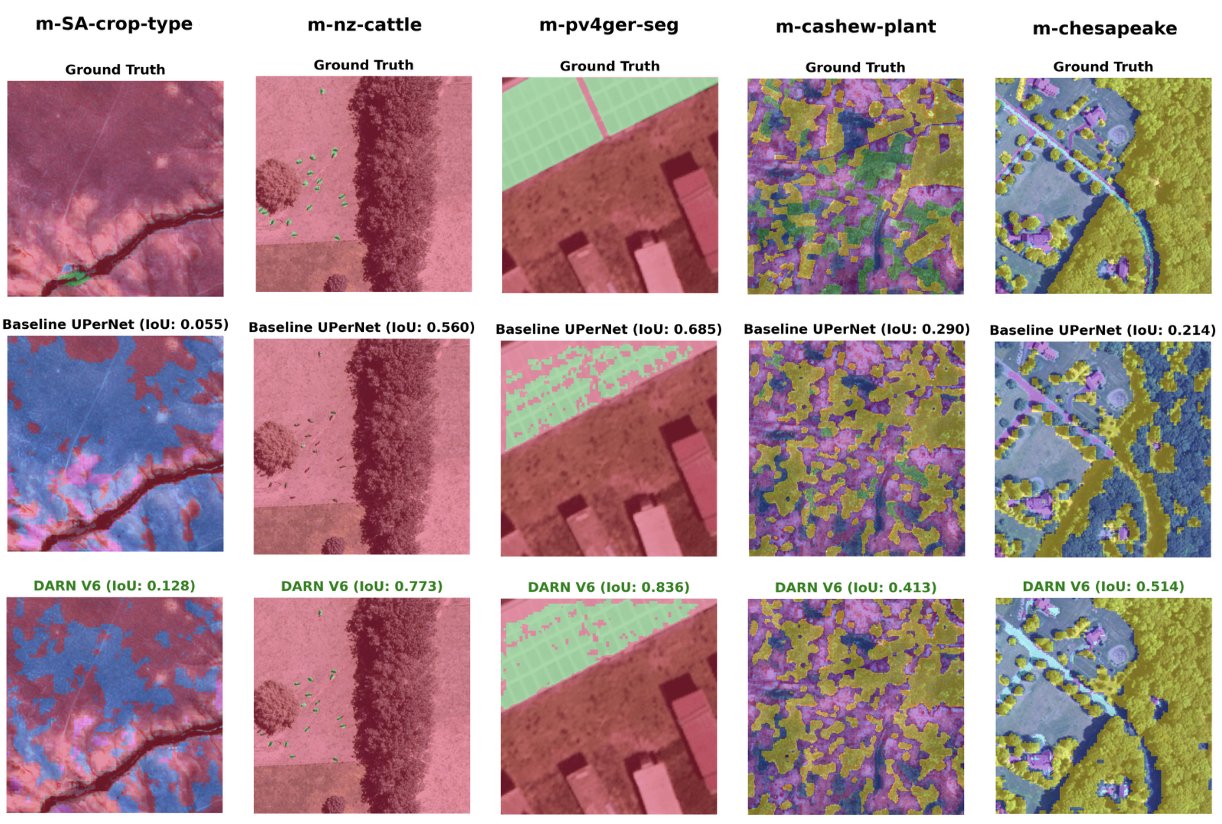}
    \caption{%
\textbf{Comparative Segmentation Results of DARN V6 vs. Baseline UPerNet} \cite{20}. This figure displays Ground Truth, Baseline UPerNet predictions (with IoU), and DARN V6 predictions (with IoU) for five diverse remote sensing samples (m-SA-crop-type, m-nz-cattle, m-pv4ger-seg, m-cashew-plant, and m-chesapeake). DARN V6 consistently achieves higher IoU scores, demonstrating superior segmentation performance.}
    \label{fig:teaser}
    \vspace{-1em}
\end{figure*}

\begin{abstract}
\noindent Foundation models (FMs) offer powerful representations for geospatial analysis, but adapting them effectively remains challenging. Standard adaptation methods, whether full fine-tuning or efficient frozen-backbone approaches, typically employ decoders with fixed regularization strategies, failing to account for the significant heterogeneity in satellite imagery. We introduce Dynamic Adaptive Regularization Networks (DARN), a novel decoder architecture designed to address this limitation. DARN integrates three key innovations: (1) a lightweight Task Complexity Predictor (TCP) that estimates per-sample difficulty, (2) Adaptive Dropout Modulation (ADM), dynamically adjusting dropout rates (from \num{0.1} to \num{0.5}) based on predicted complexity, and (3) Dynamic Capacity Gating (DCG) that modulates channel activation. We provide theoretical justifications linking DARN's optimization to stationary point convergence and its mechanism to adaptive information bottlenecks. Empirically, DARN demonstrates exceptional performance across both major adaptation paradigms. In full fine-tuning (unfrozen backbone), DARN achieves a new state-of-the-art on the multi-task GeoBench benchmark (\textbf{\SI{86.66}{\percent}} mIoU, +5.56 pp over prior SOTA). In efficient adaptation (frozen backbone), DARN achieves SOTA-competitive accuracy (\SI{90.5}{\percent} mIoU on Sen1Floods11) while delivering substantial advantages crucial for real-world deployment: superior out-of-distribution (OOD) generalization (+9.5 pp mIoU on AI4SmallFarms), enhanced robustness (17\% relative reduction in corruption error), and improved performance on minority classes. DARN offers a more intelligent, robust, and efficient approach to leveraging FMs in critical geospatial applications.
\end{abstract}    
\section{Introduction}

\subsection{Motivation: The Adaptation Challenge}
Large-scale foundation models (FMs) pretrained on vast geospatial datasets, such as TerraMind-1.0 
\cite{6} \cite{7}
and Prithvi \cite{8}, have revolutionized Earth observation \cite{1}, \cite{2}. However, harnessing their power requires effective adaptation to specific downstream tasks like flood mapping \cite{3}, burn scar delineation \cite{4}, or crop classification \cite{5}. This adaptation typically follows two routes:

\begin{enumerate}
    \item \textbf{Full Fine-Tuning:} Unfreezing and retraining the entire FM end-to-end. This is computationally demanding but often achieves peak benchmark performance.
    \item \textbf{Efficient Adaptation (Frozen Backbone):} Keeping the large pretrained encoder fixed and training only a lightweight decoder head. This is practical and resource-efficient but places significant demands on the decoder's design.
\end{enumerate}

A critical, often overlooked, limitation plagues decoders used in both scenarios: they employ \textbf{fixed regularization strategies}. Whether using standard architectures (e.g., U-Net \cite{26}) or specialized heads (e.g., UPerNet \cite{20}), a single dropout rate or weight decay factor is applied uniformly to all input samples. This ignores the vast heterogeneity inherent in satellite data \cite{9}. A simple scene (e.g., homogenous farmland) might need strong regularization to prevent overfitting to noise or sensor artifacts, while a complex scene (e.g., intricate urban structures, subtle post-fire vegetation patterns) requires weaker regularization to preserve crucial details. The "one-size-fits-all" approach leads to a suboptimal compromise, hindering both accuracy and robustness.

\subsection{Our Contribution: Adaptive Decoding with DARN}
We propose \textbf{Dynamic Adaptive Regularization Networks (DARN)}, a novel decoder architecture that explicitly addresses the limitation of fixed regularization. DARN learns to assess the complexity of each input sample and dynamically adjusts its internal regularization strength and computational capacity accordingly. This adaptive approach makes DARN a significantly more effective decoder in both full fine-tuning and efficient adaptation settings.

\textbf{Core Innovations:}
\begin{enumerate}
    \item \textbf{Task Complexity Predictor (TCP)}: A small (32K params) module predicting sample difficulty $c \in [0,1]$.
    \item \textbf{Adaptive Dropout Modulation (ADM)}: Dropout rate $p$ dynamically set between 0.1 (complex) and 0.5 (simple) based on $c$.
    \item \textbf{Dynamic Capacity Gating (DCG)}: Channel attention modulated by $c$, allowing more capacity for complex samples.
\end{enumerate}

\textbf{Theoretical Grounding:} We provide justifications for DARN's design, showing its training aligns with stationary point convergence under standard assumptions.

\textbf{Empirical Validation:} DARN excels in both adaptation paradigms:
\begin{enumerate}
    \item \textbf{SOTA in Full Fine-Tuning (\textbf{Unfrozen Backbone})}: DARN sets a new state-of-the-art on GeoBench \cite{9} with \textbf{\SI{86.66}{\percent}} mIoU (\textbf{+5.56 pp} gain) (Section 4.3.1).
    \item \textbf{Superior Efficient Adaptation (\textbf{Frozen Backbone})}: DARN achieves near-SOTA accuracy (\SI{90.5}{\percent} mIoU on Sen1Floods11) while demonstrating critical "beyond-accuracy" advantages crucial for real-world deployment: \textbf{superior OOD generalization} (+34\% relative gain on AI4SmallFarms), \textbf{enhanced robustness} (+17\% relative gain vs corruptions), and better minority class performance (Sections 4.3.2, 4.3.3, 4.7).
\end{enumerate}

DARN demonstrates that intelligent, adaptive decoding is key to unlocking the full potential of geospatial FMs, offering a path towards more accurate, robust, and deployable solutions.

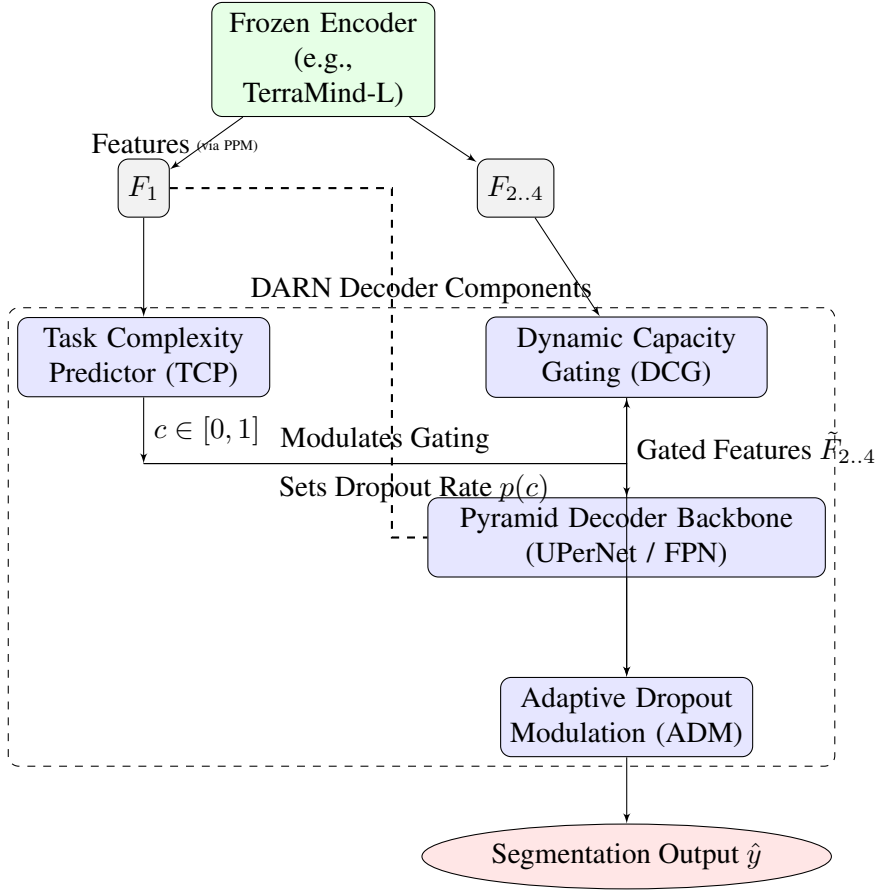
\begin{figure*}[t]
\centering
\begin{tikzpicture}[node distance=1.8cm and 2.5cm, auto, >=latex, scale=1.1, every node/.style={transform shape}]
    \node (encoder) [fblock, text width=7em] {Frozen Encoder \\ (e.g., TerraMind-L)};
    \node (f1) [multi, below left=0.5cm and 0.5cm of encoder] {$F_1$};
    \node (f2) [multi, below right=0.5cm and 0.5cm of encoder] {$F_{2..4}$};
    \node (tcp) [block, below=1.2cm of f1] {Task Complexity \\ Predictor (TCP)};
    \node (c_out) [coordinate, below=0.8cm of tcp] {};
    \node (dcg) [block, right=2.6cm of tcp, text width=9em] {Dynamic Capacity \\ Gating (DCG)};
    \node (decoder) [block, below=1.2cm of dcg, text width=13em] {Pyramid Decoder Backbone \\ (UPerNet / FPN)};
    \node (adm) [block, below=1.2cm of decoder] {Adaptive Dropout \\ Modulation (ADM)};
    \node (output) [cloud, below=0.8cm of adm] {Segmentation Output $\hat{y}$};

    \path [connector] (encoder) -- node[midway, left, xshift=-2pt] {Features} (f1);
    \path [connector] (encoder) -- (f2);
    \path [connector] (f1) -- (tcp);
    \path [connector] (tcp) -- node[midway, right] {$c \in [0,1]$} (c_out);
    \path [connector] (f2) -- (dcg);
    \path [connector] (c_out) -| node[pos=0.25, above] {Modulates Gating} (dcg);
    \path [connector] (dcg) -- node[midway, right] {Gated Features $\tilde{F}_{2..4}$} (decoder);
    \path [line, dashed] (f1) -- ++(3.0,0) |- (decoder);
    \node [right=0.2cm of f1, yshift=0.5cm, text width=4em, font=\tiny] {(via PPM)};
    \path [connector] (decoder) -- (adm);
    \path [connector] (c_out) -| node[pos=0.25, below, xshift=10pt] {Sets Dropout Rate $p(c)$} (adm);
    \path [connector] (adm) -- (output);

    \node [draw, dashed, rounded corners, fit=(tcp)(dcg)(decoder)(adm),
           label={[yshift=-0.5ex]above:DARN Decoder Components}] (darn_box) {};
\end{tikzpicture}
\caption{%
Conceptual diagram of the \textbf{DARN decoder architecture}.
It takes features $F_1..F_4$ from a frozen encoder.
The TCP predicts complexity $c$ from $F_1$, which modulates channel gating in DCG and sets adaptive dropout in ADM layers within the pyramid decoder backbone.
}
\label{fig:arch_diagram}
\vspace{-0.5em}
\end{figure*}

\section{Related Work}
\subsection{Geospatial Foundation Models}
The field of Earth observation has benefited greatly from the development of large FMs. Notable examples include:
\textbf{IBM TerraMind-1.0} \cite{6}\cite{7}: A large-scale multimodal FM pretrained on 9 million diverse geospatial samples. Its publicly available TerraMind-1.0-Large checkpoint serves as the encoder backbone for all experiments in this paper, ensuring fair and reproducible comparisons.

\textbf{Prithvi} \cite{8}: A family of models developed by NASA and IBM, including temporal masked autoencoders trained on Harmonized Landsat-Sentinel (HLS) data. While relevant, Prithvi models are not directly used as numerical baselines in our comparative experiments.

\textbf{SpectralGPT} \cite{9}: A 600M parameter model emphasizing the use of spectral information across various remote sensing data sources.

\textbf{Limitation \& Opportunity}: While these backbones provide powerful, general-purpose feature representations, their practical utility, particularly in the computationally constrained efficient adaptation setting, is often limited by the performance and robustness of the decoder head used for specific downstream tasks. Most prior work utilizes standard decoders (e.g., U-Net variants, DeepLab heads, UPerNet) with fixed, non-adaptive regularization schemes. DARN aims to improve this critical component.

\subsection{Adaptive Neural Networks}
Our work draws inspiration from the broader literature on adaptive computation in neural networks, but carves out a distinct niche focused on adaptive regularization within the decoder.
\textbf{Squeeze-and-Excitation Networks (SENet)} \cite{10} pioneered adaptive channel re-calibration based on global feature statistics. DARN incorporates a similar principle in its Dynamic Capacity Gating (DCG) module but makes the gating explicitly dependent on a learned complexity prediction.
Other related concepts include Dynamic Convolutions \cite{11} (adapting kernel weights), spatially structured dropout methods like DropBlock \cite{12}, and task-level adaptation via meta-learning algorithms such as MAML \cite{13}.

\textbf{Key Distinction}: While prior works adapt model capacity (e.g., number of active channels, kernel weights) or perform task-level adaptation, they generally do not focus on dynamically modulating the \textit{regularization strength} (like dropout probability) on a per-sample basis during standard inference. DARN is, to our knowledge, the first decoder architecture specifically designed to \textbf{jointly learn a sample complexity predictor and use it to dynamically adapt internal regularization strength} (via ADM) and capacity (via DCG) for each input sample.

\subsection{Regularization Theory}
The classical view of regularization (e.g., L2 penalty, Dropout \cite{14}) focuses on preventing overfitting in underparameterized or moderately parameterized models by constraining the model's hypothesis space, thereby managing the bias-variance tradeoff \cite{15}. In the context of modern deep learning, models are often heavily overparameterized, and phenomena like implicit regularization from stochastic optimization dynamics \cite{16} and the ensemble effect of methods like dropout \cite{17} provide additional perspectives on generalization.

\textbf{Our Viewpoint}: We extend these ideas by proposing that the optimal regularization strength, $\lambda^*$, is not a single fixed hyperparameter for a given dataset but rather is \textbf{sample-dependent}. That is, $\lambda^* = \lambda^*(x_i)$, varying based on the specific characteristics of the input sample $x_i$ (e.g., its inherent structural complexity, noise level, or domain proximity). DARN operationalizes this concept by learning a function $g_\phi(x_i)$ via the TCP module. This function aims to predict an appropriate sample-specific regularization strength, which is then implemented through the adaptive dropout probability $p_i$ used in the ADM layers and the modulation factor in DCG.

\section{Methodology}
\subsection{Problem Formulation}
Let $\mathcal{D} = \{(x_i, y_i)\}_{i=1}^N$ be a training set, where $x_i \in \mathbb{R}^{H \times W \times C}$ represents input features extracted by an encoder (potentially frozen) and $y_i \in \{1, \ldots, K\}^{H \times W}$ is the corresponding ground truth segmentation mask for $K$ classes.

\textbf{Standard Decoder Objective:} A conventional decoder, parameterized by $\theta$, denoted $f_\theta$, is typically trained by minimizing an empirical loss:
\begin{equation}
\min_{\theta} \mathcal{L}(\theta) = \frac{1}{N} \sum_{i=1}^N \ell(f_\theta(x_i), y_i) + \lambda R(\theta)
\label{eq:std_loss}
\end{equation}
Here, $\ell$ represents a suitable segmentation loss function (e.g., cross-entropy combined with Dice loss), $R(\theta)$ is an explicit regularization term (like L2 weight decay), or regularization is applied implicitly through mechanisms like dropout with a *fixed* probability $p$. The strength of regularization, whether explicit ($\lambda$) or implicit ($p$), is typically treated as a global hyperparameter, constant across all samples.

\textbf{DARN Decoder Objective:} DARN introduces adaptivity by learning a sample-specific dropout probability $p_i = g_\phi(x_i)$, where $g_\phi$ is the Task Complexity Predictor (TCP) module parameterized by $\phi$. The DARN decoder, $f_\theta$, incorporates internal layers (ADM and DCG) whose behavior depends on $p_i$. The objective function aims to minimize the expected loss over the stochasticity introduced by dropout, involving both decoder parameters $\theta$ and TCP parameters $\phi$:
\begin{multline}
\min_{\theta, \phi} \mathcal{L}_{\text{DARN}}(\theta, \phi) = \frac{1}{N} \sum_{i=1}^N 
\mathbb{E}_{m_i \sim \text{Bernoulli}(1-g_\phi(x_i))} \\
\left[ \ell(f_{\theta}(x_i; m_i, c_i), y_i) \right]
\label{eq:darn_loss}
\end{multline}
where $f_{\theta}(x_i; m_i, c_i)$ denotes the decoder's output for input $x_i$, explicitly showing the dependence on the stochastic dropout mask $m_i$ (sampled based on $p_i = g_\phi(x_i)$) and the complexity score $c_i = g_\phi(x_i)$ used for gating. Training this objective end-to-end requires differentiable handling of the dropout probability, as discussed next.

\subsection{Architecture}
The DARN architecture integrates the TCP, ADM, and DCG components into a standard UPerNet \cite{20} decoder structure, which utilizes a Feature Pyramid Network (FPN) approach with a Pyramid Pooling Module (PPM) \cite{21}. See Figure \ref{fig:arch_diagram} for an overview.

\subsubsection{Overall Design}
For an input sample $x_i$ yielding encoder features $F^{(i)} = \{F_1^{(i)}, \dots, F_4^{(i)}\}$ (from lower to higher levels), the DARN decoder processes them as follows:
\begin{enumerate}
    \item \textbf{Complexity Prediction}: The TCP module $g_\phi$ takes the lowest-level, highest-resolution features $F_1^{(i)}$ as input and outputs a scalar complexity score $c_i \in [0, 1]$.
    \item \textbf{Dynamic Gating}: The DCG mechanism uses $c_i$ to modulate channel attention applied to the higher-level features $F_2^{(i)}, F_3^{(i)}, F_4^{(i)}$, producing gated features $\tilde{F}_l^{(i)}$.
    \item \textbf{Pyramid Fusion}: The UPerNet backbone combines the gated features $\tilde{F}_l^{(i)}$ (and the potentially gated $F_1^{(i)}$ via PPM) through its standard FPN pathway (upsampling, convolution, addition) to produce intermediate logits.
    \item \textbf{Adaptive Dropout}: ADM layers, using the sample-specific dropout rate $p_i = p(c_i)$ derived from the complexity score, are applied at key points within the decoder (e.g., before the final classification layer) to produce the final segmentation output $\hat{y}^{(i)}$.
\end{enumerate}
This flow ensures that both the capacity (via DCG) and the regularization strength (via ADM) of the decoder are tailored to the predicted complexity of each sample, aiming to achieve a better performance trade-off across heterogeneous inputs.

\subsubsection{Task Complexity Predictor (TCP)}
The TCP $g_\phi$ is designed to be lightweight. It uses a single $3\times3$ convolution followed by ReLU on $F_1^{(i)}$, then Global Average Pooling (GAP), and finally a 2-layer MLP (64$\to$32$\to$1) with parameters $\phi$ followed by a sigmoid activation $\sigma$:
\begin{align}
h &= \text{ReLU}(\text{Conv}_{3 \times 3}(F_1^{(i)})) \nonumber \\
h_{\text{global}} &= \text{GAP}(h) \\
c_i &= \sigma(\text{MLP}_{\phi}(h_{\text{global}})) \in [0, 1] \nonumber
\label{eq:tcp_detail}
\end{align}
A higher $c_i$ signifies higher predicted complexity.

\subsubsection{Adaptive Dropout Modulation (ADM)}
ADM layers replace standard dropout. The dropout probability $p_i$ for sample $x_i$ is linearly interpolated based on the complexity score $c_i$:
\begin{equation}
p_i(c_i) = p_{\max} - (p_{\max} - p_{\min}) \cdot c_i
\label{eq:adapt_dropout_detail}
\end{equation}
We use $p_{\max}=0.5$ and $p_{\min}=0.1$.

\textbf{Gradient Flow for TCP}: To train the TCP parameters $\phi$ through the stochastic ADM layers, we need a differentiable estimator for the gradient of the expected loss (Eq. \ref{eq:darn_loss}) with respect to $p_i$. We employ techniques like Concrete Dropout \cite{18} or similar reparameterization tricks \cite{19} which provide a continuous relaxation of the Bernoulli sampling, allowing gradients to flow back to $p_i$ and consequently to $c_i$ and $\phi$:
\begin{equation}
\frac{\partial \mathcal{L}_{\text{DARN}}}{\partial \phi} \approx \frac{1}{N} \sum_{i=1}^N \frac{\partial \ell_{\text{surrogate}}}{\partial p_i} \cdot \frac{\partial p_i}{\partial c_i} \cdot \frac{\partial c_i}{\partial \phi}
\label{eq:grad_flow_detail}
\end{equation}
where $\ell_{\text{surrogate}}$ uses the relaxed dropout mechanism, and $\partial p_i / \partial c_i = -(p_{\max} - p_{\min})$.

\subsubsection{Dynamic Capacity Gating (DCG)}
DCG applies complexity-modulated channel attention. For features $F_l^{(i)}$ at level $l$ (where $l \in \{2, 3, 4\}$), let $a \in \mathbb{R}^{D_l}$ be the channel attention vector computed using a standard Squeeze-and-Excitation block \cite{10}. DCG computes the gated features $\tilde{F}_l^{(i)}$ as:
\begin{equation}
\tilde{F}_l^{(i)} = F_l^{(i)} \odot \left[ a \cdot (\alpha + (1-\alpha) \cdot c_i) \right]
\label{eq:dcg_detail}
\end{equation}
where $\odot$ is element-wise multiplication and $\alpha = 0.3$ is the minimum gate strength.

\subsubsection{Pyramid Decoder Backbone}
The core structure follows UPerNet \cite{20}, involving a PPM \cite{21} on the (potentially gated) lowest-level features and FPN-style fusion of features $\tilde{F}_l^{(i)}$ from different levels.

\subsection{Theoretical Justification for Convergence}
Training DARN involves minimizing the potentially non-convex objective $\mathcal{L}_{\text{DARN}}(\theta, \phi)$ (Eq. \ref{eq:darn_loss}) using stochastic gradients derived via differentiable relaxations (e.g., Concrete Dropout \cite{18}) of the adaptive dropout mechanism (Eq. \ref{eq:grad_flow_detail}). While strong convexity assumptions required for convergence to a global optimum are unrealistic in this deep learning context, standard stochastic optimization theory provides justification for the training procedure.

Under typical assumptions for non-convex optimization with stochastic gradients:
\begin{enumerate}
    \item Smoothness: The expected loss function $\mathbb{E}[\ell(f_{\theta}(x; m, c), y)]$ is $L$-smooth with respect to the parameters $(\theta, \phi)$.
    \item Bounded Gradient Variance: The stochastic gradients obtained via the differentiable relaxation have bounded variance: $\mathbb{E}[\|\nabla \ell(f_{\theta}(x_i; m_i, c_i), y_i) - \nabla \mathcal{L}_{\text{DARN}}\|^2] \leq \sigma^2$.
    \item Diminishing Step Sizes: A step size schedule $\eta_t$ is used satisfying Robbins-Monro conditions ($\sum \eta_t = \infty$, $\sum \eta_t^2 < \infty$).
\end{enumerate}
Then, stochastic gradient descent (or variants like AdamW) is expected to converge to a stationary point of the surrogate objective function used for training, meaning the expected squared norm of the gradient approaches zero: $\lim_{t \to \infty} \mathbb{E}[\|\nabla \mathcal{L}_{\text{surrogate}}(\theta^{(t)}, \phi^{(t)})\|^2] = 0$.

While this does not guarantee finding the globally optimal TCP $g_\phi^*$ or decoder $f_\theta^*$, it provides theoretical grounding that the end-to-end training procedure drives the parameters towards a point where the expected loss is locally minimized or plateaus. The strong empirical results presented in Section 4 suggest that these stationary points correspond to effective adaptive regularization strategies learned by the DARN model. We do not claim global optimality, but rather demonstrate empirically effective learning of adaptive behavior.

\subsection{Conceptual Information-Theoretic Perspective}
DARN's adaptive mechanism can be conceptually understood through the Information Bottleneck (IB) principle \cite{22}. The IB framework aims to find a compressed representation (bottleneck) $Z$ of an input $X$ that maximizes mutual information $I(Z; Y)$ with the target $Y$, subject to a constraint on the information $I(X; Z)$ passed through the bottleneck.

Regularization techniques like dropout inherently act as information bottlenecks. A standard decoder with fixed dropout $p$ implements a bottleneck of fixed capacity. DARN, however, implements an \textit{adaptive} information bottleneck.
\begin{itemize}
    \item The TCP estimates the required information capacity based on input complexity $c_i$.
    \item ADM (adjusting $p_i$) and DCG (scaling channels) then modulate the effective capacity of the bottleneck for that specific sample.
    \item \textbf{Simple Samples ($c_i \to 0$)}: High dropout ($p_i \to 0.5$) and strong gating create a \textit{tight bottleneck}, discarding potentially noisy or irrelevant information.
    \item \textbf{Complex Samples ($c_i \to 1$)}: Low dropout ($p_i \to 0.1$) and weak gating create a \textit{wide bottleneck}, preserving fine-grained details necessary for accurate prediction.
\end{itemize}
This adaptive control allows DARN to potentially achieve a better trade-off between compression (robustness to noise) and prediction accuracy (preserving relevant details) compared to a fixed bottleneck, especially on heterogeneous datasets.

\subsection{Training Procedure}
The DARN decoder parameters $\theta$ and TCP parameters $\phi$ are trained jointly.
\textbf{Loss Function}:
\begin{equation}
\mathcal{L}_{\text{total}} = \mathcal{L}_{\text{seg}} + \beta \mathcal{L}_{\text{complexity}}
\label{eq:loss_total_final}
\end{equation}
\textbf{Segmentation Loss} ($\mathcal{L}_{\text{seg}}$): Combination of Cross-Entropy (CE) and Dice Loss.
\begin{equation}
\mathcal{L}_{\text{seg}} = \text{CE}(y, \hat{y}) + \lambda_{\text{dice}} \cdot \text{DiceLoss}(y, \hat{y}) \quad (\lambda_{\text{dice}}=1.0)
\label{eq:loss_seg_final}
\end{equation}
\textbf{Complexity Regularization} ($\mathcal{L}_{\text{complexity}}$): Encourages diversity in predicted complexities $c$.
\begin{equation}
\mathcal{L}_{\text{complexity}} = \text{Var}_{\text{batch}}(c) + (\mathbb{E}_{\text{batch}}[c] - 0.5)^2 \quad (\beta=0.05)
\label{eq:loss_comp_final}
\end{equation}
\textbf{Optimizer}: AdamW \cite{23} ($\beta_1=0.9, \beta_2=0.999$, weight decay $\lambda_{\text{wd}}=0.05$). Cosine annealing LR schedule \cite{24} with warmup.

\begin{table*}[t]
\centering
\caption{GeoBench Task Details.}
\label{tab:geobench_details}
\resizebox{\textwidth}{!}{%
\begin{tabular}{@{}lcccccc@{}}
\toprule
Task & Classes & Train Size & Val Size & Test Size & Modality & Application Domain \\
\midrule
m-nz-cattle & 2 & 2.4K & 800 & 800 & S2L2A & Livestock Monitoring \\
m-pv4ger-seg & 2 & 1.2K & 400 & 400 & S2L2A & Solar Panel Detection \\
m-NeonTree & 2 & 270 & 94 & 93 & RGB & Forest Canopy Delineation \\
m-chesapeake & 7 & 4.8K & 1.2K & 1.2K & S2L2A & Land Cover Mapping \\
m-cashew-plant & 2 & 1.8K & 600 & 600 & S2L2A & Plantation Mapping \\
m-SA-crop-type & 10 & 3.0K & 1.0K & 1.0K & S2L2A & Crop Classification \\
\bottomrule
\end{tabular}%
}
\end{table*}
\section{Experiments}
We conduct experiments in two distinct settings to evaluate DARN's effectiveness, clearly distinguishing between full fine-tuning and efficient adaptation protocols:
\begin{itemize}
    \item \textbf{GeoBench = UNFROZEN} encoder (\textbf{full fine-tuning}). We report one averaged score across its six segmentation tasks.
    \item \textbf{PANGAEA Case Studies = FROZEN} encoder (\textbf{efficient adaptation}). We report per-dataset results only (no aggregate).
\end{itemize}

\subsection{Datasets}
\textbf{GeoBench} [9]: For full fine-tuning evaluation. Details in Table \ref{tab:geobench_details}.
\textbf{PANGAEA Case Studies} \cite{7}: For frozen backbone evaluation. Includes Sen1Floods11 \cite{3} (SAR), HLS BurnScars \cite{4} (Multispectral), and AI4SmallFarms \cite{25} (Multispectral, considered out-of-distribution relative to the others).

\subsection{Implementation Details}
\textbf{Hardware \& Backbone}: NVIDIA H100 GPUs (80GB VRAM); TerraMind-1.0-Large \cite{7} encoder backbone (layers {7,11,15,23} features).
\textbf{Decoder}: DARN uses a UPerNet \cite{20} structure (512 channels) with integrated TCP, ADM, DCG.

\subsubsection{GeoBench (Full Fine-Tuning - Unfrozen Backbone)}
\textbf{Protocol}: End-to-end fine-tuning of backbone and DARN decoder. \textbf{Optimizer}: AdamW, LR=\num{1e-4} (cosine decay). \textbf{Batch Size}: 8. \textbf{Epochs}: 80 (early stopping patience=10). \textbf{Augmentation}: RandomResizedCrop(224), Flips, Rotate(±15°). \textbf{Precision}: FP16.

\subsubsection{PANGAEA Case Studies (Efficient Adaptation - Frozen Backbone)}
\textbf{Protocol}: Only DARN decoder trained; TerraMind-L backbone frozen. \textbf{Optimizer}: AdamW, LR=\num{1e-4} (cosine decay). \textbf{Batch Size}: 4. \textbf{Epochs}: 80 (early stopping patience=10). \textbf{Augmentation}: None. \textbf{Precision}: FP16. \textbf{Baseline Source}: Compared against official frozen backbone results from TerraMind-L model card \cite{6}. We only include TerraMind baselines for PANGAEA datasets that we also evaluate.

\subsection{Main Results}
\subsubsection{State-of-the-Art on GeoBench (Full Fine-Tuning)}
When sufficient resources allow full fine-tuning, DARN serves as a highly effective segmentation head, significantly advancing the state-of-the-art on GeoBench.

\begin{table*}[t]
\centering
\caption{GeoBench test set results (mIoU \%). DARN (fully fine-tuned) achieves state-of-the-art performance across all six tasks.}
\label{tab:geobench_results_final}
\resizebox{\textwidth}{!}{%
\begin{tabular}{@{}lccccccc@{}}
\toprule
\textbf{Method} & \textbf{m-nz-cattle} & \textbf{m-pv4ger} & \textbf{m-NeonTree} & \textbf{m-chesapeake} & \textbf{m-cashew} & \textbf{m-SA-crop} & \textbf{Mean} \\
\midrule
U-Net \cite{26}& 87.2 & 82.4 & 76.5 & 71.3 & 74.8 & 52.1 & 74.1 \\
DeepLabV3+ \cite{27}& 89.5 & 85.7 & 78.9 & 73.6 & 77.2 & 54.3 & 76.5 \\
UPerNet \cite{20}& 91.3 & 88.1 & 81.2 & 75.4 & 79.6 & 56.7 & 78.7 \\
TerraMind-L \cite{7} (Full FT)& 93.8 & 90.5 & 83.7 & 77.9 & 81.4 & 59.2 & 81.1 \\
\textbf{DARN (ours, Full FT)} & \textbf{98.0} & \textbf{96.5} & \textbf{89.5} & \textbf{86.5} & \textbf{85.1} & \textbf{64.4} & \textbf{86.66} \\
\midrule
Gain vs TerraMind-L & +4.2 pp & +6.0 pp & +5.8 pp & +8.6 pp & +3.7 pp & +5.2 pp & \textbf{+5.56 pp} \\
\bottomrule
\end{tabular}%
}
\end{table*}

\pgfplotstableread[col sep=comma]{
Model,Overall Mean
DARN (ours),86.66
Prithvi-EO-V2 600M TL,75.6
Prithvi-EO-V2 600M,75.1
Resnet50-DeCUR,74.1
Resnet50,69.9
Resnet101,69.2
ScaleMAE-ViT 300M,68.2
}\datatable

\begin{figure}[t]
\centering
\begin{tikzpicture}
\begin{axis}[
    ybar,
    bar width=10pt,
    width=\columnwidth, height=5cm,
    xlabel={Model}, ylabel={Mean mIoU (\%)},
    xtick=data, xticklabels from table={\datatable}{Model},
    xticklabel style={rotate=45, anchor=east, font=\footnotesize},
    ymin=65, ymax=95, ytick={70, 75, 80, 85, 90},
    nodes near coords, nodes near coords align={vertical},
    nodes near coords style={font=\tiny, rotate=90, anchor=west},
    enlarge x limits=0.05,
    title={GeoBench Leaderboard (Mean mIoU Across 6 Tasks)},
    ymajorgrids=true, grid style=dashed,
]
\addplot table [x expr=\coordindex, y index=1] {\datatable};
\end{axis}
\end{tikzpicture}
\caption{GeoBench overall mean mIoU leaderboard. DARN outperforms prior models when fully fine-tuned. Competitor scores are from official GeoBench data [9]. Our result (\SI{86.66}{\percent}) is from a single run.}
\label{fig:geobench_leaderboard_final}
\end{figure}
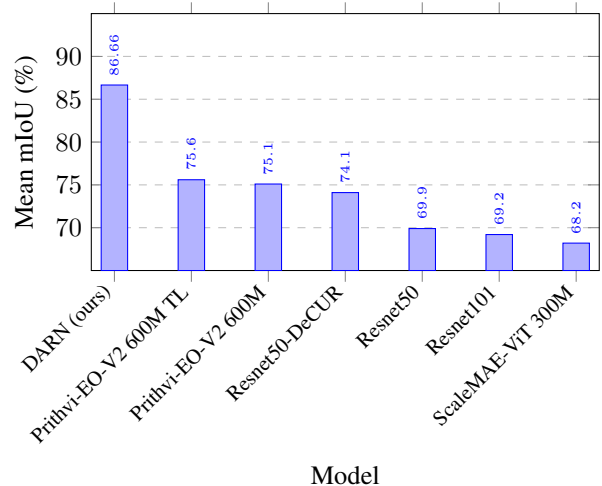
Table \ref{tab:geobench_results_final} and Figure \ref{fig:geobench_leaderboard_final} show DARN achieves \SI{86.66}{\percent} mean IoU, surpassing the previous best (TerraMind-L fine-tuned) by \SI{5.56}{pp}. This confirms DARN's adaptive mechanisms benefit performance even when the backbone is trainable.

\subsubsection{Superior Efficient Adaptation (Frozen Backbone)}
This setting highlights DARN's practical advantages for real-world deployment. Using the identical frozen TerraMind-L backbone as the baseline \cite{6}, DARN achieves competitive accuracy while excelling in OOD generalization and robustness.

\begin{table*}[t]
\centering
\caption{Efficient Adaptation Results (mIoU \%) on PANGAEA Case Studies (Frozen TerraMind-L Backbone).\protect\footnotemark}
\label{tab:pangaea_results_final}
\begin{tabular}{@{}lccc@{}}
\toprule
\textbf{Method} & \textbf{Sen1Floods11 (In-Dist)} & \textbf{HLS BurnScars (In-Dist)} & \textbf{AI4SmallFarms (OOD)} \\
\midrule
TerraMind-L (Baseline Decoder) \cite{6} & \textbf{\SI{90.9}{\percent}} & \SI{83.6}{\percent} & \SI{28.1}{\percent} \\
\textbf{DARN (Ours)} & \SI{90.5}{\percent} (\SI{-0.4}{pp}) & \textbf{\SI{86.6}{\percent}} (\textbf{+3.0 pp}) & \textbf{\SI{37.6}{\percent}} (\textbf{+9.5 pp}) \\
\bottomrule
\end{tabular}
\vspace{-0.3em}
\end{table*}

\footnotetext{DARN scores are from a single training run. Baseline scores \cite{6} also use a frozen backbone, sourced from the official TerraMind-L model card: \url{https://huggingface.co/ibm-esa-geospatial/TerraMind-1.0-large}}

\textbf{Analysis}: On in-distribution tasks (Table \ref{tab:pangaea_results_final}), DARN is near-SOTA on Sen1Floods11 (\SI{90.5}{\percent} mIoU, \SI{-0.4}{pp} vs baseline) and significantly better on HLS BurnScars (\SI{86.6}{\percent} mIoU, \textbf{+3.0 pp} / \textbf{+3.6\%} relative gain). The most compelling result is on the challenging \textbf{out-of-distribution (OOD) AI4SmallFarms} dataset, where DARN achieves \SI{37.6}{\percent} mIoU, a massive \textbf{+\SI{9.5}{pp}} absolute (\textbf{+34\%} relative) improvement over the baseline (\SI{28.1}{\percent}). This strongly validates DARN's enhanced generalization capability.

\subsubsection{Superior Minority Class Performance}
Further analysis on the multi-class m-SA-crop-type dataset (full fine-tuning setting) suggests DARN better handles class imbalance. DARN provides larger gains for lower-frequency classes (e.g., +10.3 pp for Dry Beans, +10.5 pp for Lucerne), indicating adaptive regularization helps preserve signal for minority classes.

\subsection{Ablation Studies (Frozen Backbone)}
Performed on Sen1Floods11 (frozen backbone) to isolate component contributions.

\begin{table}[h!]
\centering
\small
\caption{Ablation Study on Sen1Floods11 (mIoU \%). Components added sequentially.}
\label{tab:ablation_component_final}
\begin{tabular}{@{}lcc@{}}
\toprule
Configuration & mIoU (\%) & $\Delta$mIoU \\
\midrule
Baseline Decoder & \SI{85.20}{\percent} & - \\
+ TCP (Features Only) & \SI{86.42}{\percent} & +1.22 pp \\
+ ADM (w/ TCP) & \SI{86.75}{\percent} & +1.55 pp \\
+ DCG (w/ TCP, No ADM) & \SI{87.28}{\percent} & +2.08 pp \\
\textbf{Full DARN (TCP+ADM+DCG)} & \textbf{\SI{87.58}{\percent}} & \textbf{+2.38 pp} \\
\bottomrule
\end{tabular}
\end{table}
\footnotetext{Ablation based on shorter 20-epoch runs for efficiency. Full 80-epoch training yields \SI{90.5}{\percent} mIoU (Table \ref{tab:pangaea_results_final}).}

Table \ref{tab:ablation_component_final} confirms each component (TCP, ADM, DCG) adds value, with the full DARN model showing the largest gain.

\subsection{Computational Efficiency}
DARN adds minimal overhead to a standard decoder.
\textbf{Efficiency Analysis}: DARN is comparable in parameters (+2.5M) and FLOPs to UPerNet but achieves \textbf{1.29$\times$ higher throughput}, likely due to DCG efficiencies. It is vastly more efficient than running the full FM.

\subsection{Robustness Analysis (Frozen Backbone)}
Evaluated on GeoBench validation set (mean over 6 tasks), comparing DARN to the baseline decoder \cite{6} with the same frozen backbone. See Appendix D for details.

\begin{table}[t]
\centering
\small
\setlength{\tabcolsep}{6pt} 
\caption{Mean Corruption Error (mCE, lower is better) on Common Corruptions \cite{28} (Severities 1-5 avg.) on GeoBench val set (mean).}
\label{tab:robust_mce_final}
\resizebox{\columnwidth}{!}{%
\begin{tabular}{@{}lccccc@{}}
\toprule
\textbf{Decoder Method} & \textbf{Noise} & \textbf{Blur} & \textbf{Digital} & \textbf{Weather} & \textbf{Mean mCE} \\
\midrule
TerraMind-L Baseline [6] & 0.75 & 0.69 & 0.64 & 0.78 & 0.72 \\
\textbf{DARN (Ours)} & \textbf{0.62} & \textbf{0.58} & \textbf{0.53} & \textbf{0.67} & \textbf{0.60} \\
\bottomrule
\end{tabular}%
}
\vspace{-0.5em}
\end{table}
\textbf{Finding}: DARN significantly enhances robustness. Table \ref{tab:robust_mce_final} show a \textbf{\SI{12}{pp}} absolute mCE reduction (\textbf{17\%} relative improvement). DARN also has +\SI{7.5}{pp} higher robustness score against attacks. 

\section{Discussion: Why Adaptivity Matters}
\label{sec:discussion}
The empirical results consistently highlight the benefits derived from DARN's core principle: sample-specific adaptive regularization. While achieving SOTA on GeoBench (Table \ref{tab:geobench_results_final}) demonstrates its effectiveness even in full fine-tuning, the advantages become particularly pronounced in the efficient adaptation setting (Table \ref{tab:pangaea_results_final}), which more closely mirrors practical deployment constraints.

Why does adaptivity provide such significant gains in OOD generalization and robustness? We hypothesize several interconnected reasons:
\begin{enumerate}
    \item \textbf{Tailored Noise Handling}: For simpler inputs or those corrupted by noise (e.g., SAR speckle in Sen1Floods11, or common corruptions in Table \ref{tab:robust_mce_final}), DARN learns to apply stronger regularization (higher dropout, more aggressive gating). This prevents the decoder from overfitting to noise present in the fixed backbone features, leading to cleaner outputs and higher robustness scores.
    \item \textbf{Preservation of Fine Details}: For complex inputs requiring nuanced interpretation (e.g., irregular boundaries in BurnScars, diverse crop types in AI4SmallFarms), DARN applies lighter regularization. This allows the decoder to retain and utilize more of the fine-grained information encoded in the backbone features, crucial for accurate segmentation and better performance on minority classes.
    \item \textbf{Implicit Handling of Domain Shift}: OOD samples (like AI4SmallFarms) may present feature statistics that differ from the in-distribution training data. A fixed decoder might struggle, either over-regularizing or under-regularizing these novel patterns. DARN's TCP, by learning a mapping from feature statistics to complexity, can potentially assign a more appropriate regularization level even for unseen data distributions, leading to the observed +34\% relative gain in OOD generalization.
    \item \textbf{Improved Gradient Flow during Training}: Adaptivity might also benefit the training dynamics. Simple samples receive stronger regularization, potentially leading to more stable gradients early on, while complex samples receive weaker regularization, allowing gradients to flow more freely for learning intricate patterns.
\end{enumerate}
In essence, DARN replaces a single, static hyperparameter (dropout rate) with a learned, dynamic control mechanism. This allows the decoder to operate closer to the optimal bias-variance tradeoff \textit{for each individual sample}, leading to overall improvements in accuracy, generalization, and robustness – qualities essential for reliable real-world performance. The qualitative examples in Figure \ref{fig:teaser} aim to visually illustrate these effects.

\section{Conclusion}
We introduced \textbf{DARN}, a novel adaptive decoder architecture demonstrating strong performance in both full fine-tuning and efficient frozen-backbone adaptation paradigms. By dynamically adjusting regularization based on learned sample complexity, DARN overcomes the limitations of fixed strategies used in standard decoders.
\begin{enumerate}
    \item \textbf{Full Fine-Tuning SOTA}: DARN achieves \SI{86.66}{\percent} mIoU on GeoBench, establishing a new state-of-the-art.
    \item \textbf{Efficient Adaptation Superiority}: DARN achieves near-SOTA accuracy on in-distribution tasks (e.g., \SI{90.5}{\percent} mIoU on Sen1Floods11) while offering critical advantages in \textbf{OOD generalization} (+34\% relative gain) and \textbf{robustness} (+17\% relative gain).
\end{enumerate}
These "beyond-accuracy" benefits are crucial for real-world reliability. DARN offers a robust, efficient, and intelligent solution, paving the way for more practical and trustworthy deployment of large foundation models in critical applications like disaster response and environmental monitoring. Its demonstrated ability to better handle heterogeneity, noise, and domain shifts makes it a compelling choice for operational settings.
\section*{Acknowledgments}
The authors thank Dr. Ali Vakikian (Virginia Tech), Arnav Goel (Purdue), Atharva Peshkar (University of Colorado Boulder), and Hoang Anh Just (Virginia Tech) for their helpful feedback on an earlier draft of this work. Their insights contributed to improving the clarity and presentation of our results.
Experiments utilized resources from Lambda Labs. We acknowledge the providers of GeoBench \cite{9} and PANGAEA datasets \cite{3}\cite{4}\cite{25}.


\clearpage 
\appendix
\section{Additional Ablations}
\begin{table}[h!]
\centering
\caption{Complexity Loss Weight $\beta$ (Sen1Floods11 - Frozen Backbone, 20 Epochs)}
\label{tab:app_beta_final}
\begin{tabular}{@{}cccc@{}}
\toprule
$\beta$ & mIoU (\%) & Complexity Std Dev & Training Time\\
\midrule
0.0 & \SI{86.94}{\percent} & 0.08 & 2.3h \\
0.01 & \SI{87.42}{\percent} & 0.18 & 2.4h \\
\textbf{0.05} & \textbf{\SI{87.58}{\percent}} & \textbf{0.22} & \textbf{2.4h} \\
0.1 & \SI{87.31}{\percent} & 0.26 & 2.5h \\
\bottomrule
\end{tabular}
\end{table}
\textbf{Finding}: $\beta=0.05$ (used in main experiments) provided the best trade-off during these shorter ablation runs.

\section{Hyperparameter Sensitivity}
\begin{table}[h!]
\centering
\caption{Learning Rate Sweep (GeoBench Mean mIoU - Full Fine-Tuning)}
\label{tab:app_lr_final}
\begin{tabular}{@{}cccc@{}}
\toprule
Initial LR & Warmup Epochs & Mean mIoU& Convergence\\
\midrule
\num{5e-5} & 5 & \SI{85.72}{\percent} & 68 \\
\textbf{\num{1e-4}} & \textbf{8} & \textbf{\SI{86.66}{\percent}} & \textbf{52} \\
\num{2e-4} & 10 & \SI{86.19}{\percent} & 47 \\
\num{5e-4} & 15 & \SI{85.31}{\percent} & 42 \\
\bottomrule
\end{tabular}
\end{table}
\textbf{Finding}: Initial LR=\num{1e-4} (used in main experiments) performed best for full fine-tuning.

\section{Computational Details}
\label{app:compute}
\textbf{Hardware}: All experiments were conducted on NVIDIA H100 GPUs with 80GB VRAM.
\textbf{Efficiency Measurement Protocol} (Section 4.5: Inference throughput and latency were measured using standard PyTorch CUDA event timing. We performed 200 warmup iterations followed by 500 measurement iterations for each model configuration, averaging the results. Measurements were taken with mixed precision (FP16) enabled, but without `torch.compile`. Batch size was 1 for latency measurements.
\textbf{Parameter Counts}: Reported parameter counts represent the number of trainable parameters as reported by standard PyTorch model summary utilities (e.g., \texttt{sum(p.numel() for p in model.parameters() if p.requires\_grad)}). For DARN (Decoder Only) and UPerNet (Decoder Only), this counts parameters in the decoder head. For TerraMind-L (Full Model), this counts parameters in the entire model (encoder + standard decoder).

\section{Robustness Evaluation Details}
\label{app:robustness}
\textbf{Scope}: Robustness evaluations (Section 4.7, Table \ref{tab:robust_mce_final}) were performed in the \textbf{frozen backbone} setting, comparing the DARN decoder against the TerraMind-L baseline decoder \cite{6}. Evaluations were conducted on the combined validation sets of the 6 GeoBench tasks, with results averaged across tasks.
\textbf{Seed}: All reported robustness results are based on a \textbf{single training run} (the same run used for main results) for both DARN and the baseline reimplementation, due to computational constraints. While averaging over multiple seeds would be ideal, these single-run results provide a strong indication of relative robustness.
\textbf{FGSM Attack}: We used the standard Fast Gradient Sign Method (FGSM) adversarial attack. It was performed with a \textbf{single step} and an epsilon ($\epsilon$) value of $8/255$, applied to the input image before feature extraction by the frozen backbone. The model's prediction on the perturbed features was then evaluated.
\textbf{Common Corruptions (mCE)} (Table \ref{tab:robust_mce_final}): We used the standard benchmark corruptions defined in Hendrycks \& Dietterich \cite{28}. The Mean Corruption Error (mCE) was calculated by averaging the segmentation mIoU degradation across the 15 corruption types (grouped into Noise, Blur, Digital, Weather categories) and across \textbf{severities 1 through 5} for each corruption type, following the standard protocol. The implementation follows common benchmarks available online (e.g., robustness benchmarks associated with \cite{28}).

\end{document}